\documentclass[10pt,twocolumn,letterpaper]{article}

\usepackage[pagenumbers]{cvpr} 

\usepackage{placeins}

\definecolor{iccvblue}{rgb}{0.21,0.49,0.74}
\usepackage[pagebackref,breaklinks,colorlinks,allcolors=iccvblue]{hyperref}

\def\paperID{*****} 
\def\confName{CVPR}
\def\confYear{2026}

\title{SmokeGS-R: Physics-Guided Pseudo-Clean 3DGS for\\ Real-World Multi-View Smoke Restoration}

\author{
Xueming Fu$^{1}$ \quad
Lixia Han$^{2}$ \\ \\
$^{1}$\,University of Science and Technology of China \quad
$^{2}$\,Nanjing University of Aeronautics and Astronautics
}

\begin{document}
\maketitle

\begin{abstract}
Real-world smoke simultaneously attenuates scene radiance, adds airlight, and destabilizes multi-view appearance consistency, making robust 3D reconstruction particularly difficult. We present \textbf{SmokeGS-R}, a practical pipeline developed for the NTIRE 2026 3D Restoration and Reconstruction Track 2 challenge. The key idea is to decouple geometry recovery from appearance correction: we generate physics-guided pseudo-clean supervision with a refined dark channel prior and guided filtering, train a sharp clean-only 3D Gaussian Splatting source model, and then harmonize its renderings with a donor ensemble using geometric-mean reference aggregation, LAB-space Reinhard transfer, and light Gaussian smoothing. On the official challenge testing leaderboard, the final submission achieved \mbox{PSNR $=15.217$} and \mbox{SSIM $=0.666$}. After the public release of RealX3D, we re-evaluated the same frozen result on the seven released challenge scenes without retraining and obtained \mbox{PSNR $=15.209$}, \mbox{SSIM $=0.644$}, and \mbox{LPIPS $=0.551$}, outperforming the strongest official baseline average on the same scenes by $+3.68$ dB PSNR. These results suggest that a geometry-first reconstruction strategy combined with stable post-render appearance harmonization is an effective recipe for real-world multi-view smoke restoration. The code is available at \url{https://github.com/windrise/3drr_Track2_SmokeGS-R}.
\end{abstract}

\section{Introduction}
\label{sec:intro}

Reliable multi-view reconstruction in the presence of smoke remains difficult even for modern neural scene representations. In real indoor environments, smoke simultaneously attenuates direct radiance, introduces additive airlight, and changes appearance across viewpoints in a depth-dependent manner. These effects degrade both geometry estimation and novel-view rendering, making standard reconstruction assumptions unreliable. The recently released RealX3D benchmark explicitly highlights this gap by showing that real smoke scenes remain challenging for existing scattering-aware methods and even for carefully engineered 3D reconstruction pipelines~\cite{liu2025realx3d}.

This problem became especially concrete in the NTIRE 2026 3D Restoration and Reconstruction (3DRR) Track 2 challenge~\cite{liu2026ntire}, where the task is to remove smoke from multi-view observations and synthesize restored novel views. The challenge also stimulated a broader wave of contemporaneous reports on smoke reconstruction and related adverse-condition restoration problems. During the challenge, we observed a recurring failure mode: methods that aggressively model smoke inside the 3D representation often distort the underlying scene geometry, while methods that produce sharper structure still suffer from residual color cast and airlight bias. This tension motivated us to design a solution that separates \emph{geometry preservation} from \emph{appearance correction}.

Our final challenge method, \textbf{SmokeGS-R}, follows this design principle. First, we generate physics-guided pseudo-clean supervision from smoky training images using a refined dark channel prior (DCP) pipeline. Second, we train a sharp clean-only 3D Gaussian Splatting (3DGS) source branch~\cite{kerbl20233d} to preserve scene structure. Third, we use several complementary donor branches to provide stable appearance statistics, and harmonize the sharp source rendering in CIE-LAB space through a geometric-mean reference and Reinhard color transfer. This simple decoupled design proved more stable than directly forcing a single model to jointly solve geometry recovery, smoke disentanglement, and appearance restoration.

With the public release of RealX3D on Hugging Face, the challenge solution can now be reorganized into a benchmark-oriented short paper. The present manuscript is aligned with our challenge report and arXiv version of SmokeGS-R~\cite{fu2026smokegs}. In this work, we keep the challenge-proven method intact, summarize the method formally, position it against the official RealX3D smoke baselines, and directly re-evaluate the frozen final result on the released smoke scenes. Our best official challenge submission achieved PSNR $15.217$ and SSIM $0.666$ on the testing leaderboard, and the same frozen result reaches PSNR $15.209$ and SSIM $0.644$ on the seven released challenge scenes of the public RealX3D package.

The contributions of this report are three-fold:
\begin{itemize}
    \item We present \textbf{SmokeGS-R}, a physics-guided pseudo-clean 3DGS pipeline that decouples geometry recovery from smoke-induced appearance correction.
    \item We show how a \textbf{clean-only sharp source branch} and a \textbf{multi-reference donor ensemble} can be combined through LAB-space harmonization for practical smoke restoration.
    \item We connect the challenge solution to the \textbf{public RealX3D smoke benchmark}, audit the released official baselines, and show that the frozen final result transfers strongly to the released scenes without additional retraining.
\end{itemize}

\section{Related Work}
\label{sec:related_work}

Real-world degraded 3D reconstruction has recently attracted increasing attention, but benchmark coverage has long lagged behind method development. RealX3D~\cite{liu2025realx3d} is particularly important because it provides pixel-aligned low-quality and reference views, quarter-resolution data for challenge-style evaluation, and a dedicated smoke subset with both training-view and NVS metrics. For the present report, RealX3D serves not only as a benchmark but also as the public bridge that lets us re-evaluate a frozen challenge solution under a released and reproducible protocol.

\paragraph{Scattering-aware reconstruction.}
Several methods explicitly model participating media inside neural rendering or 3D reconstruction pipelines. SeaThru-NeRF~\cite{levy2023seathru} introduces physically grounded radiance-field modeling in scattering media, while I2-NeRF~\cite{liu2025i2nerf}, Watersplatting~\cite{li2025watersplatting}, and SeaSplat~\cite{yang2025seasplat} extend similar principles to more complex media interactions and explicit 3D representations. More recent challenge-time smoke reports further explore multimodal-prior-guided reconstruction, generative multi-stage synthesis, and dehaze-then-splat strategies~\cite{zheng20263d,cao2026gensmoke,chen2026dehaze}. These methods are highly relevant because smoke and underwater scattering share attenuation and in-scattering components. However, the RealX3D study shows that strong synthetic or domain-specific formulations do not directly translate to real smoke scenes~\cite{liu2025realx3d}. This observation closely matches our challenge experience: aggressive internal smoke modeling can easily entangle medium appearance with scene geometry.

\paragraph{Robust 3DGS under degraded observations.}
Vanilla 3DGS~\cite{kerbl20233d} provides an efficient and high-fidelity reconstruction backbone, but it assumes reasonably stable view-consistent observations. Under adverse conditions, recent methods introduce view-adaptive priors, photometric corrections, or restoration-aware optimization. Examples include Luminance-GS~\cite{cui2025luminance} and LITA-GS~\cite{zhou2025lita}, which target challenging illumination conditions, as well as more recent low-light 3DGS variants based on dual-branch enhancement, staged reliability modeling, and progressive pruning~\cite{liu2026elog,zhu2026naka,guo2026reliability}. These directions complement the broader robust reconstruction frameworks surveyed in RealX3D~\cite{liu2025realx3d}. Our method differs from these lines in one practical aspect: rather than integrating every correction into a single monolithic 3D model, we retain a geometry-first source branch and postpone part of the appearance restoration to a controlled test-time harmonization stage.

\paragraph{Physics-guided restoration priors.}
For smoke removal and haze removal, classical image formation priors remain surprisingly effective when used as stable pseudo-clean generators. The dark channel prior~\cite{he2011dark} provides a simple way to estimate transmission from corrupted observations, and guided filtering~\cite{he2013guided} helps refine the recovered pseudo-clean targets without introducing brittle learned artifacts. In addition, global color transfer methods such as Reinhard transfer~\cite{reinhard2001color} offer a lightweight mechanism for matching low-frequency appearance statistics. Our challenge solution combines these ideas with 3DGS: DCP-based pseudo-clean supervision shapes the geometry-first source branch, and LAB-space Reinhard transfer harmonizes appearance after rendering.

\paragraph{Concurrent challenge context.}
The NTIRE 2026 RealX3D challenge also appeared alongside a broader family of adverse-condition restoration reports~\cite{liu2026ntire}. Although several of these works target adjacent 2D problems rather than multi-view smoke reconstruction, they highlight complementary ideas such as training-free ensembling, dual-branch specialization, data-centric self-ensemble, and augmentation-driven robustness~\cite{chang2026training,ge2026dual,chang2026beyond,ge2026clip}. Relative to these contemporaneous studies, our method stays deliberately lightweight and focuses on a geometry-first smoke reconstruction pipeline with post-render appearance harmonization rather than a heavier integrated restoration model.

\section{Methods}
\label{subsec:methods}

\subsection{Overview}

\begin{figure*}[t]
    \centering
    \includegraphics[width=0.92\textwidth]{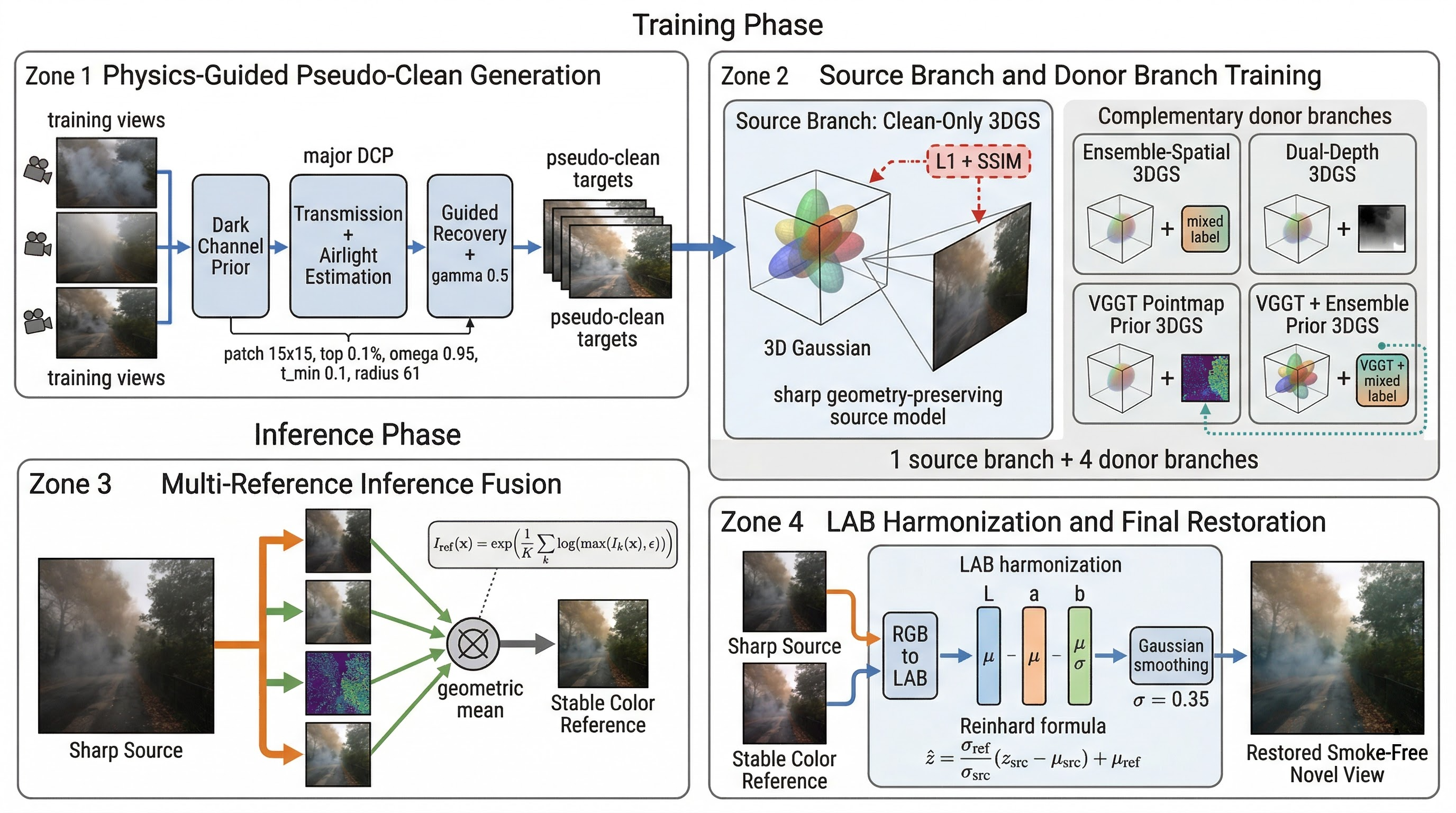}
    \caption{Overview of SmokeGS-R. A refined DCP inversion pipeline produces physics-guided pseudo-clean supervision for a clean-only source 3DGS, while four complementary donor branches provide auxiliary appearance priors. At inference, a five-render geometric-mean reference is constructed and used to harmonize the sharp source rendering in LAB space, followed by light Gaussian smoothing for final smoke-free novel-view restoration.}
    \label{fig:pipeline}
\end{figure*}

Figure~\ref{fig:pipeline} summarizes the proposed pipeline. SmokeGS-R contains three stages: (i) physics-guided pseudo-clean generation from smoky inputs, (ii) a geometry-first clean-only 3DGS source model, and (iii) multi-reference appearance harmonization using donor renderings. The design deliberately avoids forcing a single branch to jointly explain scene geometry, medium attenuation, and airlight bias. Instead, we keep the geometry branch sharp and use donor branches only as stable appearance references.

\subsection{Physics-Guided Pseudo-Clean Supervision}

We begin from the standard atmospheric scattering model
\begin{equation}
I_{\text{obs}}(x) = I_{\text{clean}}(x)\, t(x) + A \,(1 - t(x)),
\end{equation}
where $t(x)$ is the transmission map and $A$ is global airlight. Following the dark channel prior (DCP)~\cite{he2011dark}, we compute
\begin{equation}
I_{\text{dark}}(x)=\min_{c\in\{R,G,B\}} \min_{y\in\Omega(x)} I^c(y),
\end{equation}
where $\Omega(x)$ is a local $15\times15$ patch. The initial transmission estimate is
\begin{equation}
\hat{t}(x)=1-\omega \min_{y\in\Omega(x)} \min_c \frac{I^c(y)}{A^c},
\end{equation}
with $\omega = 0.95$. We then refine $\hat{t}(x)$ using guided filtering~\cite{he2013guided} with radius $61$ and $\epsilon=10^{-3}$, and recover the pseudo-clean image as
\begin{equation}
\hat{I}_{\text{clean}}(x)=\frac{I(x)-A}{\max(\hat{t}(x), t_{\min})}+A,
\end{equation}
with $t_{\min}=0.1$. Finally, we apply gamma enhancement with $\gamma=0.5$ to compensate for the low-contrast appearance commonly observed after inversion. These pseudo-clean targets are not treated as perfect ground truth; instead, they serve as stable geometric supervision that is substantially cleaner than the original smoky inputs.

\subsection{Geometry-First Source Branch and Donor Ensemble}

The sharp source model is a clean-only 3DGS branch trained for 5{,}000 iterations using the standard photometric loss
\begin{equation}
\mathcal{L}_{\text{photo}}=(1-\lambda_{\text{ssim}})\mathcal{L}_{1}
+\lambda_{\text{ssim}}(1-\text{SSIM}),
\end{equation}
with $\lambda_{\text{ssim}}=0.2$. For the source branch, additional appearance-heavy auxiliary constraints are intentionally avoided so that the model focuses on preserving clean structure, object boundaries, and stable multi-view geometry.

Alongside the source branch, we maintain four donor branches that encode complementary priors:
\begin{itemize}
    \item an ensemble-spatial branch,
    \item a dual-depth branch,
    \item a VGGT-prior branch,
    \item a VGGT-ensemble-prior branch.
\end{itemize}
These donors are not used to replace the source geometry. Instead, they provide a diverse but stable appearance pool under different smoke densities and color casts. This design was chosen after extensive challenge experimentation showed that donor renderings are often more useful as appearance references than as direct geometry replacements.

\subsection{LAB-Space Multi-Reference Harmonization}

At inference time, we first render the sharp source image $I_{\text{src}}$. We then render four donor images and compute a five-render geometric-mean reference
\begin{equation}
I_{\text{ref}}(x)=\exp\!\left(\frac{1}{K}\sum_{k=1}^{K}
\log(\max(I_k(x),\varepsilon))\right),
\end{equation}
with $K=5$. The geometric mean suppresses unstable outlier renderings while preserving globally consistent low-frequency appearance statistics.

We next perform Reinhard color transfer~\cite{reinhard2001color} in CIE-LAB space. For each channel $c\in\{L,a,b\}$,
\begin{equation}
\hat{z}_c = \frac{\sigma_{\text{ref}}^{(c)}}{\sigma_{\text{src}}^{(c)}}
\left(z_{\text{src}}^{(c)}-\mu_{\text{src}}^{(c)}\right)+\mu_{\text{ref}}^{(c)}.
\end{equation}
This operation matches the first- and second-order appearance statistics of the source rendering to those of the ensemble reference. The LAB parameterization is useful because luminance and chromatic components are more cleanly separated than in RGB, which lets us correct smoke-induced color bias while preserving the local geometry already captured by the source branch. We further apply a light Gaussian smoothing with $\sigma=0.35$ to suppress residual splatting artifacts and then encode the final image with conservative JPEG settings (quality $95$, 4:2:0 subsampling).

\subsection{Design Rationale}

The key empirical lesson behind SmokeGS-R is that robust smoke restoration is not only a matter of stronger physical modeling. In the challenge, several more aggressive decomposition attempts were less stable than the final source-plus-harmonization design. We therefore favor physically interpretable but globally stable operators: DCP for pseudo-clean initialization, clean-only 3DGS for geometry, and low-frequency ensemble statistics for appearance. This keeps the pipeline simple enough to be reproducible while still addressing the dominant color and visibility degradation caused by smoke.

\section{Experiment}
\label{subsec:experiment}

\subsection{Benchmark Setup}

We focus on the publicly released smoke subset of RealX3D~\cite{liu2025realx3d}. The Hugging Face release provides two image resolutions: the original full-resolution package in \texttt{data/} and the quarter-resolution package in \texttt{data\_4/}. The latter is particularly important because it matches the resolution regime used in the NTIRE 2026 3DRR challenge and contains the same eight smoke scenes used throughout our challenge study: Akikaze, Futaba, Hinoki, Koharu, Midori, Natsume, Shirohana, and Tsubaki. The release also includes official baseline renderings in \texttt{baseline\_results/} together with per-scene \texttt{eval\_train.json} and \texttt{eval\_test.json} files. In practice, the released \texttt{smoke/<scene>/train|val|test} folders serve as clean reference views, while the official method outputs are stored separately under \texttt{baseline\_results/}. This structure allows direct post-hoc comparison between our frozen challenge outputs and the released reference test views.

Following RealX3D, the primary photometric metrics are PSNR, SSIM, and LPIPS. In the released public smoke package, we directly observe the reference RGB images, camera metadata, official baseline renderings, and the organizers' precomputed evaluation JSON files. We therefore use the public release in three complementary ways: (1) to summarize the official public smoke baselines exactly as released, (2) to recompute public baseline averages over the subset of scenes covered by our final challenge package, and (3) to directly compare our frozen final result against the released clean reference test views without retraining.

\subsection{Official Smoke Baselines on RealX3D}

Table~\ref{tab:realx3d-smoke-baselines} summarizes the official NVS results reported for smoke in RealX3D Table~2. Two observations are especially relevant for our method design. First, the absolute performance is still low across all methods, which confirms that real smoke remains difficult even for dedicated scattering-aware pipelines. Second, stronger physical modeling does not automatically improve robustness under real smoke; the baseline numbers remain tightly clustered and none of the official methods fully resolves the geometry--appearance entanglement problem.

\begin{table*}[t]
    \centering
    \small
    \setlength{\tabcolsep}{8pt}
    \begin{tabular}{lccc}
        \toprule
        Method & NVS PSNR$\uparrow$ & NVS SSIM$\uparrow$ & NVS LPIPS$\downarrow$ \\
        \midrule
        3DGS~\cite{kerbl20233d} & 9.76 & 0.499 & 0.659 \\
        SeaThru-NeRF~\cite{levy2023seathru} & 7.55 & 0.464 & 0.683 \\
        Watersplatting~\cite{li2025watersplatting} & 10.78 & 0.445 & 0.723 \\
        SeaSplat~\cite{yang2025seasplat} & 10.42 & 0.446 & 0.774 \\
        I2-NeRF~\cite{liu2025i2nerf} & 8.40 & 0.283 & 0.699 \\
        \bottomrule
    \end{tabular}
    \caption{Official NVS smoke results reported by RealX3D~\cite{liu2025realx3d}. These are the benchmark-level averages reported in the dataset paper.}
    \label{tab:realx3d-smoke-baselines}
\end{table*}

The official discussion in RealX3D further notes that real smoke is harder than synthetic scattering settings because it exhibits spatially varying density, non-uniform airlight, and strong view-dependent attenuation~\cite{liu2025realx3d}. This is exactly the regime in which we found challenge-time color harmonization to be useful: direct scene reconstruction and smoke appearance correction should not be forced into a single unstable representation.

\subsection{Public Release Audit}

To ground the paper in the released assets rather than only the PDF summary table, we also parsed the official \texttt{eval\_test.json} files from all eight public smoke scenes. The resulting averages are shown in Table~\ref{tab:public-smoke-baselines}. They are consistent with the overall benchmark message: performance remains low under real smoke, and the strongest method depends heavily on the scene. Interestingly, the plain 3DGS baseline is the strongest average model on this exact released subset, which further supports our geometry-first design choice.

\begin{table}[t]
    \centering
    \small
    \begin{tabular}{lccc}
        \toprule
        Method & PSNR$\uparrow$ & SSIM$\uparrow$ & LPIPS$\downarrow$ \\
        \midrule
        3DGS & 11.439 & 0.579 & 0.631 \\
        SeaThru-NeRF & 9.013 & 0.547 & 0.651 \\
        WaterSplatting & 9.302 & 0.426 & 0.713 \\
        SeaSplat & 9.277 & 0.445 & 0.760 \\
        I2-NeRF & 7.138 & 0.255 & 0.704 \\
        \bottomrule
    \end{tabular}
    \caption{Averages recomputed from the official public \texttt{eval\_test.json} files over the eight released smoke scenes.}
    \label{tab:public-smoke-baselines}
\end{table}

\subsection{Challenge Evidence}

Our direct quantitative evidence comes from the official NTIRE 2026 3DRR Track 2 leaderboard, where the quarter-resolution protocol matches the released \texttt{data\_4} smoke setting. Table~\ref{tab:challenge-results} reports the two frozen results that define the final method version used in this report.

\begin{table}[h]
    \centering
    \small
    \begin{tabular}{lcc}
        \toprule
        Setting & PSNR$\uparrow$ & SSIM$\uparrow$ \\
        \midrule
        Development best (\#624877) & 15.682 & 0.669 \\
        Testing best (\#635505) & 15.217 & 0.666 \\
        \bottomrule
    \end{tabular}
    \caption{Official challenge results for the final SmokeGS-R pipeline. The development score corresponds to our best development-side submission, while the testing score corresponds to the final valid result package used for release.}
    \label{tab:challenge-results}
\end{table}

The final challenge submission used a frozen geometry-first source branch together with four donor branches and LAB-space appearance harmonization. This result is particularly important because it was not obtained by training a more complicated smoke decomposition network. Instead, it came from a practical decoupling strategy: preserve geometry in the source branch and correct appearance after rendering using stable ensemble statistics. In the accompanying source release, the final testing artifact is frozen as \texttt{test\_ct\_g035.zip}, making the public release audit fully reproducible from the final code package.

\subsection{Direct Public Comparison on the Released Challenge Scenes}

Because the public release contains the clean reference test views for the former challenge scenes, we can directly compare our frozen final result package against the official public baselines on the same seven scenes contained in \texttt{Final\_result.zip}. Table~\ref{tab:public-direct-comparison} reports this comparison over $28$ released test views. SmokeGS-R reaches $15.209$ PSNR, $0.644$ SSIM, and $0.551$ LPIPS, which is $+3.68$ dB above the strongest single official baseline average (3DGS at $11.530$ PSNR) and still $+2.35$ dB above a scene-wise oracle that selects the best official baseline separately for each scene. In LPIPS, our result also improves on the strongest single public baseline by $0.076$ on average. The gain is especially large on Koharu, Midori, Natsume, and Shirohana, while Tsubaki remains the only scene where plain 3DGS is slightly stronger in PSNR.

\begin{table}[t]
    \centering
    \small
    \begin{tabular}{lccc}
        \toprule
        Method & PSNR$\uparrow$ & SSIM$\uparrow$ & LPIPS$\downarrow$ \\
        \midrule
        3DGS & 11.530 & 0.583 & 0.627 \\
        SeaThru-NeRF & 9.133 & 0.552 & 0.648 \\
        WaterSplatting & 9.268 & 0.424 & 0.708 \\
        SeaSplat & 9.001 & 0.438 & 0.745 \\
        I2-NeRF & 7.130 & 0.256 & 0.699 \\
        \midrule
        SmokeGS-R (ours) & \textbf{15.209} & \textbf{0.644} & \textbf{0.551} \\
        \bottomrule
    \end{tabular}
    \caption{Direct comparison on the seven released challenge scenes (28 public test views) covered by our final result package. Official baselines use the organizers' released \texttt{eval\_test.json} values; our row is recomputed against the released reference views. Akikaze is excluded here because it is not part of the frozen final testing package.}
    \label{tab:public-direct-comparison}
\end{table}

\begin{figure}[t]
    \centering
    \includegraphics[width=\columnwidth]{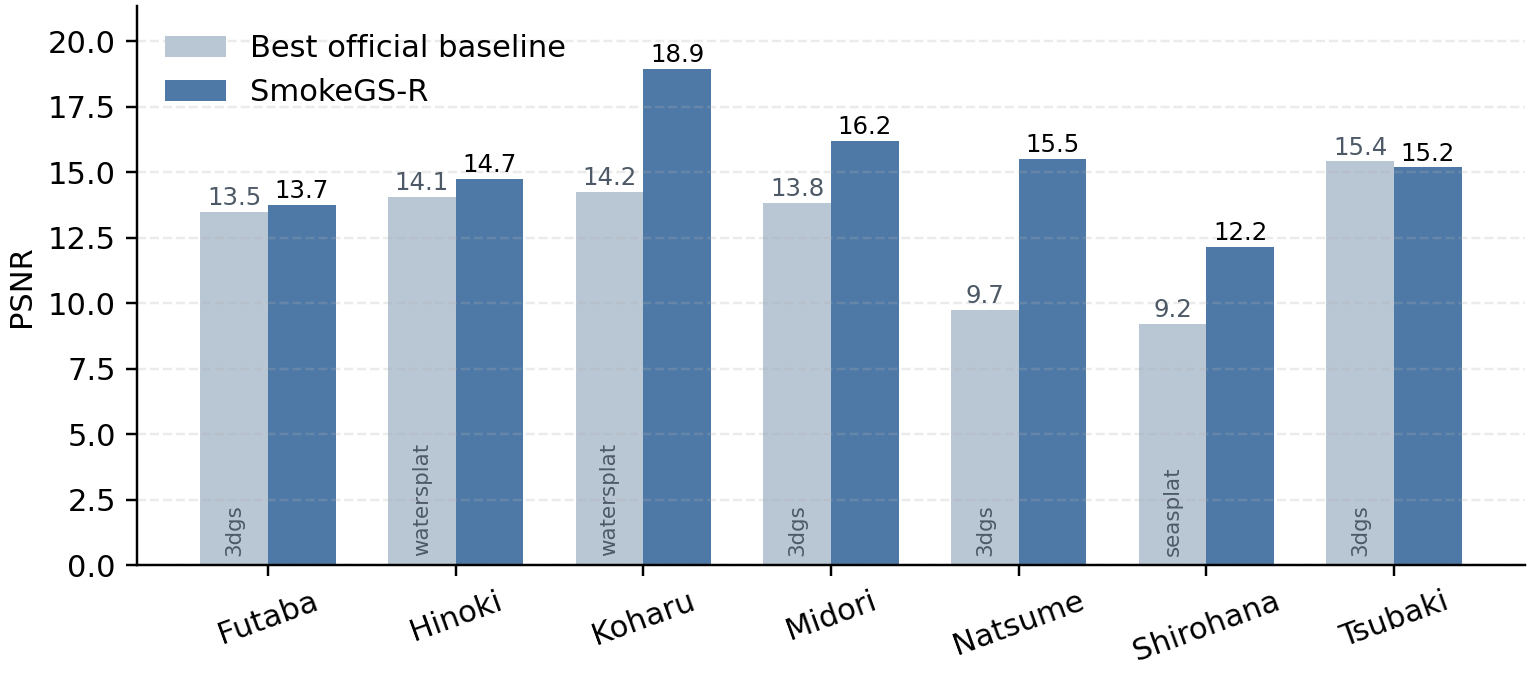}
    \caption{Scene-wise PSNR on the seven released challenge scenes. For each scene, we compare SmokeGS-R with the strongest official public baseline on that scene.}
    \label{fig:public-scene-psnr}
\end{figure}

Figure~\ref{fig:public-scene-psnr} further breaks the public comparison down by scene. The pattern is revealing: once geometry is preserved, appearance harmonization contributes the largest gains precisely in the scenes with the strongest residual smoke veil, while the benefit becomes marginal in the relatively clean Tsubaki scene. This scene-dependent behavior is consistent with our challenge-time diagnosis that smoke restoration errors are dominated less by a single universal failure mode than by how strongly medium appearance leaks into the reconstructed scene colors.

\subsection{Qualitative Transfer on the Released Smoke Scenes}

\begin{figure*}[htbp]
    \centering
    \includegraphics[width=\textwidth]{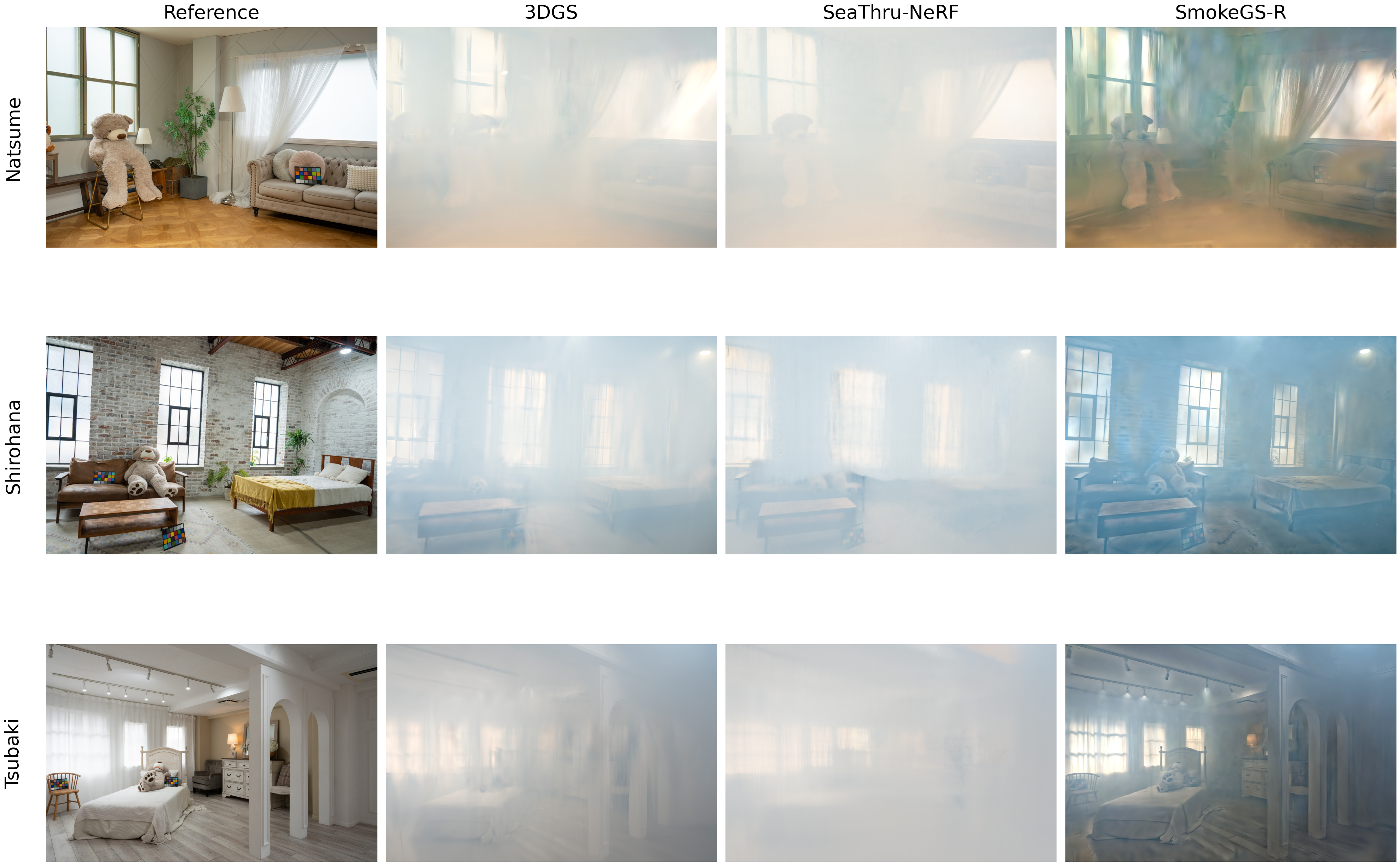}
    \caption{Qualitative comparison on the publicly released smoke scenes. Each row corresponds to one scene from the former testing split. We show the released reference view, the official 3DGS baseline, the official SeaThru-NeRF baseline, and our final frozen SmokeGS-R result.}
    \label{fig:public-smoke-qualitative}
\end{figure*}

Figure~\ref{fig:public-smoke-qualitative} shows the released reference test view together with two official public baselines and our final frozen result on the three former testing scenes. The trend is stable across all inspected views: our pipeline preserves the object contours and background structure of the sharp source branch while reducing the dominant smoke veil more aggressively than the official baselines. The benefit is especially visible in the bear silhouette and nearby tabletop boundaries, where naive reconstruction tends to retain heavy veiling or lose local contrast.

This report intentionally preserves the challenge-proven method and figure design while reconnecting them to the released public assets. The resulting picture is consistent across sources: official public baselines remain far from solving real smoke, and the strongest version of our method is the simple decoupled one, namely a sharp clean-only source 3DGS followed by stable multi-reference appearance harmonization rather than an aggressively entangled internal smoke model.

\FloatBarrier
\section{Conclusion}
\label{sec:conclusion}

We presented SmokeGS-R, a practical multi-view smoke restoration pipeline developed for the NTIRE 2026 3DRR Track 2 challenge and reorganized here as a benchmark-oriented short paper. The method combines physics-guided pseudo-clean supervision, a geometry-first clean-only 3DGS source branch, and LAB-space appearance harmonization from a complementary donor ensemble. The core lesson is simple: in real smoke scenes, geometry recovery and appearance correction are better handled in a decoupled way than by an overly aggressive single-branch decomposition.

The public release of RealX3D makes it possible to verify this conclusion beyond the original leaderboard. On the seven released challenge scenes covered by our final frozen package, SmokeGS-R reaches $15.209$ PSNR, $0.644$ SSIM, and $0.551$ LPIPS, clearly outperforming the released official baselines on average. This result suggests that a geometry-first reconstruction followed by stable appearance harmonization is a strong and practical recipe for real smoke restoration, even without resorting to a heavily entangled internal smoke decomposition model.

A limitation of the current pipeline is that its gain is scene-dependent: the improvement is largest in scenes with strong residual smoke veil, while relatively clean scenes such as Tsubaki leave less room for post-render harmonization. Nevertheless, the method is fully reproducible from the released code and frozen result package, and the public RealX3D release now provides a concrete benchmark for studying when such decoupled restoration strategies are most effective.

\section*{Acknowledgments}
We thank the organizers of the NTIRE 2026 3D Restoration and Reconstruction Challenge and the authors of RealX3D for releasing the public benchmark, baseline renderings, and geometry assets. We also acknowledge the open-source projects that supported this work, including 3D Gaussian Splatting, gsplat, VGGT, and DehazeFormer.

{
    \small
    \bibliographystyle{ieeenat_fullname}
    \bibliography{main}
}

\end{document}